\documentclass[11pt,a4paper]{article}
\usepackage[hyperref]{emnlp2020}
\usepackage{times}
\usepackage{latexsym}

\usepackage{soul}
\usepackage{graphicx}
\usepackage{amsmath}
\usepackage{url}
\usepackage{booktabs}
\usepackage{tabularx}
\usepackage{array,multirow}
\usepackage{subcaption}
\usepackage{float}
\usepackage{makecell}
\usepackage{framed}
\usepackage{enumitem}
\usepackage{titlesec} 
\usepackage{microtype}
\usepackage{siunitx}
\usepackage{verbatim}
\usepackage{background}

\usepackage{microtype}

\definecolor{mygray}{gray}{0.7}

\usepackage{amsfonts, amsthm, bm, amssymb}

\newcommand{\plus}[1]{{\color{royalblue}{#1}}}
\newcommand{\minus}[1]{{\color{orangered}{#1}}}
\newcommand{\regex}[1]{\texttt{r"#1"}}

\newcommand{\lex}[1]{\textit{#1}}
\newcolumntype{C}{>{\centering\arraybackslash}X}

\definecolor{royalblue}{HTML}{2058DC}
\definecolor{orangered}{HTML}{E34132}
\definecolor{hlcolor}{rgb}{.9, 1, .4}
\sethlcolor{hlcolor}

\colorlet{shadecolor}{hlcolor}

\newcommand{\nostar}{({~~~~~~})}
\newcommand{\onestar}{({~~*~~})}
\newcommand{\twostars}{({~**~})}
\newcommand{\threestars}{({***})}

\newcommand{\STAB}[1]{\begin{tabular}{@{}c@{}}#1\end{tabular}}
\newcommand{\rotate}[2]{\multirow{#1}{*}{\STAB{\rotatebox[origin=c]{90}{#2}}}}

\aclfinalcopy 

\title{Detecting Attackable Sentences in Arguments}

\author{Yohan Jo$^1$ ~ Seojin Bang$^1$ ~ Emaad Manzoor$^2$ ~ Eduard Hovy$^1$ ~ Chris Reed$^3$ \\
  $^1$School of Computer Science, Carnegie Mellon University, USA \\
  $^2$Heinz College of Information Systems and Public Policy, Carnegie Mellon University, USA \\
  $^3$Centre for Argument Technology, University of Dundee, UK \\
  \texttt{\{yohanj,seojinb\}@cs.cmu.edu}, \texttt{\{emaad,hovy\}@cmu.edu} \\  \texttt{c.a.reed@dundee.ac.uk} \\}

\date{}

\backgroundsetup{
    angle=0,
    scale=1,
    color=blue,
    position=current page.south,
    vshift=1.5cm,
    contents={\scriptsize \ifnum\value{page}=1 \shortstack[l]{@inproceedings\{Jo:2020attack, \\
    \hspace{2cm}author = \{Jo, Yohan and Bang, Seojin and Manzoor, Emaad and Hovy, Eduard and Reed, Chris \}, \\
    \hspace{2cm}title = \{Detecting Attackable Sentences in Arguments\}, \\
    \hspace{2cm}booktitle = \{Proceedings of the 2020 Conference on Empirical Methods in Natural Language Processing\}, year = \{2020\}\}} \fi}
}

\begin{document}
\maketitle
\begin{abstract}
    Finding attackable sentences in an argument is the first step toward successful refutation in argumentation. We present a first large-scale analysis of sentence attackability in online arguments. We analyze driving reasons for attacks in argumentation and identify relevant characteristics of sentences. We demonstrate that a sentence's attackability is associated with many of these characteristics regarding the sentence's content, proposition types, and tone, and that an external knowledge source can provide useful information about attackability. Building on these findings, we demonstrate that machine learning models can automatically detect attackable sentences in arguments, significantly better than several baselines and comparably well to laypeople.\footnote{Our data and source code are available at: \url{github.com/yohanjo/emnlp20_arg_attack}}
\end{abstract}

\section{Introduction}

\begin{figure}[t]
    \centering
    \includegraphics[width=.79\linewidth]{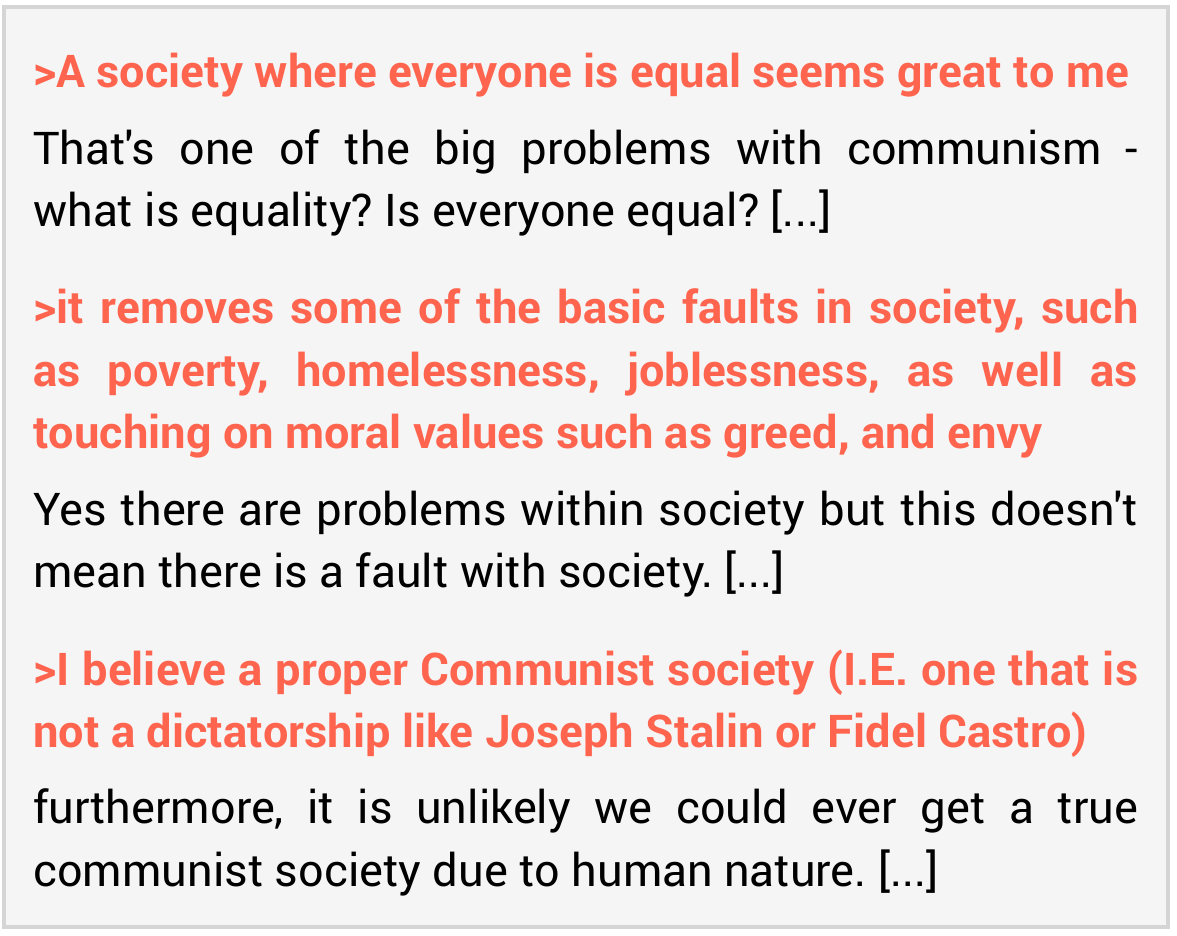}
    \caption{A comment to a post entitled ``I believe that Communism is not as bad as everyone says''. It quotes and attacks some sentences in the post (red with ``{\scriptsize\textgreater}'')}
    \label{fig:quote_example}
\end{figure}

Effectively refuting an argument is an important skill in persuasion dialogue, and the first step is to find appropriate points to attack in the argument. Prior work in NLP has studied argument quality~\cite{Wachsmuth:2017uy,Habernal:2016um} and counterargument generation~\cite{Hua:2019un,Wachsmuth:2018retr}. But these studies mainly concern an argument's overall quality and making counterarguments toward the main claim, without investigating what parts of an argument are attackable for successful persuasion. Nevertheless, attacking specific points of an argument is common and effective; in our data of online discussions, challengers who successfully change the original poster's view are 1.5 times more likely to quote specific sentences of the argument for attacks than unsuccessful challengers (Figure \ref{fig:quote_example}). In this paper, we examine how to computationally detect attackable sentences in arguments. This attackability information would help people make persuasive refutations and strengthen an argument by solidifying potentially attackable points.


To examine the characteristics of attackable sentences in an argument, we first conduct a qualitative analysis of reasons for attacks in online arguments. Our data comes from discussions in the ChangeMyView (CMV) forum on Reddit. In CMV, users challenge the viewpoints of original posters (OPs), and those who succeed receive a $\Delta$ from the OPs. In this setting, sentences that are attacked and lead to the OP's view change are considered ``attackable'', i.e., targets that are worth attacking. Admittedly, persuasion has to do with ``how'' to attack as well, but this is beyond the scope of this paper.
We only focus on choosing proper sentences to attack, which is a prerequisite for effective persuasion.

This analysis of reasons for attacks, along with argumentation theory and discourse studies, provide insights into what characteristics of sentences are relevant to attackability. Informed by these insights, we extract features that represent relevant sentence characteristics, clustered into four categories: content, external knowledge, proposition types, and tone. We demonstrate the effects of individual features on sentence attackability, in regard to whether a sentence would be \emph{attacked} and whether a sentence would be attacked \emph{successfully}.

Building on these findings, we examine the efficacy of machine learning models in detecting attackable sentences in arguments. We demonstrate that their decisions match the gold standard significantly better than several baselines and comparably well to laypeople.

To the best of our knowledge, this work is the first large-scale analysis of sentence attackability in arguments. Our contributions are as follows:
\begin{itemize}[topsep=0pt,itemsep=-5pt]
    \item We introduce the problem of detecting attackable sentences in arguments and release the processed data from online discussions and the external knowledge source we used.
    \item We analyze driving reasons for attacks in arguments and the effects of sentence characteristics on a sentence's attackability.
    \item We demonstrate the performance of machine learning models for detecting attackable sentences, setting a baseline for this challenging task and suggesting future directions.  
\end{itemize}

\section{Background}

The strength of an argument is a long-studied topic, dating back to \newcite{Aristotle:2007va}, who suggested three aspects of argument persuasiveness: ethos (the arguer's credibility), logos (logic), and pathos (appeal to the hearer's emotion). More recently, \newcite{Wachsmuth:2017ty} summarized various aspects of argument quality studied in argumentation theory and NLP, such as clarity, relevance, and arrangement. Some research took empirical approaches and collected argument evaluation criteria from human evaluators~\cite{Habernal:2016um,Wachsmuth:2017uy}. By adopting some of these aspects, computational models have been proposed to automatically evaluate argument quality in various settings, such as essays~\cite{Ke:2019gv}, online comments~\cite{Gu:2018if}, and pairwise ranking~\cite{Habernal:2016tg}.
While these taxonomies help understand and evaluate the quality of an argument as a whole, little empirical analysis has been done in terms of what to attack in an argument to persuade the arguer.

What can be attacked in an argument has been studied more in argumentation theory. Particularly, \newcite{Walton:2008schem} present argumentation schemes and critical questions (CQs). Argument schemes are reasoning types commonly used in daily argumentation. 
For instance, the scheme of \emph{argument from cause to effect} has the conclusion ``\textit{B} will occur'' supported by the premise ``if \textit{A} occurs, \textit{B} will occur. In this case, \text{A} occurs''.
Each scheme is associated with a set of CQs for judging the argument to be good or fallacious. 
CQs for the above scheme include ``How strong is the causal generalization?'' and ``Are there other factors that interfere with the causal effect?''
Unlike the general argument quality described in the previous paragraph, CQs serve as an evaluation tool that specify local attackable points in an argument. They have been adopted for grading essays~\cite{Song:2017cn} and teaching argumentation skills~\cite{Nussbaum:2018ce}. Some of the sentence characteristics in our work are informed by argumentation schemes and CQs.

NLP researchers have widely studied the effectiveness of counterarguments on persuasion~\cite{Tan:2016bk,CanoBasave:2016uq,Wei:2016ui,Wang:2017wu,Morio:2019cj} and how to generate counterarguments~\cite{Hua:2019un,Wachsmuth:2018retr}. Most of the work focuses on the characteristics of counterarguments with respect to topics and styles, without consideration of what points to attack. On the other hand, some studies aimed to model the salience of individual sentences in attacked arguments by paying different degrees of attention to sentences using attention mechanism~\cite{Jo:2018tta,Ji:2018wm}. While their approaches helped to predict the success of persuasion, it was difficult to interpret what constitute the salience or attackability of sentences. To address this limitation, we quantify and analyze the characteristics of sentences that are attacked and lead to the arguer's view change.

\section{Data}
Here we describe how we collected and labeled our data.

\subsection{Data Collection}
We use online discussions from the ChangeMyView (CMV) subreddit\footnote{\url{https://www.reddit.com/r/changemyview}}. In this forum, users post their views on various issues and invite other users to challenge their views. If a comment changes the original poster (OP)'s view, the OP acknowledges it by replying to the comment with a $\Delta$ symbol. The high quality of the discussions in this forum is maintained through several moderation rules, such as the minimum length of an original post and the maximum response time of OPs. As a result, CMV discussions have been used in many NLP studies~\cite{Chakrabarty:2019gv,Morio:2019cj,Jo:2018tta,Musi:2017hs,Wei:2016ui,Tan:2016bk}. 

We scraped CMV posts and comments written between January 1, 2014 and September 30, 2019, using the Pushshift API. We split them into a dev set (Jan 2014--Jan 2018 for training and Feb 2018--Nov 2018 for validation) and a test set (Dec 2018--Sep 2019), with the ratio of 6:2:2. We split the data by time to measure our models' generality to unseen subjects.

As the characteristics of arguments vary across different issues, we categorized the posts into domains using LDA. 
For each post, we chose as its domain the topic that has the highest standard score; topics comprising common words were excluded. We tried different numbers of topics (25, 30, 35, 40) and finalized on 40, as it achieves the lowest perplexity. This process resulted in 30 domains (excluding common-word topics): media, abortion, sex, election, Reddit, human economy, gender, race, family, life, crime, relationship, movie, world, game, tax, law, money, drug, war, religion, job, food, power, school, college, music, gun, and Jewish (from most frequent to least, ranging 5\%--2\%).


\subsection{Labeling Attackability}
Since we are interested in which parts of a post are attacked by comments and whether the attacks lead to successful view changes, our goal here is to label each sentence in a post as \emph{successfully attacked}, \emph{unsuccessfully attacked}, or \emph{unattacked}. We only consider comments directly replying to each post (top-level comments), as lower-level comments usually address the same points as their parent comments (as will be validated at the end of the section).

\paragraph{Attacked vs. Unattacked:} Some comments use direct quotes with the $>$ symbol to address specific sentences of the post (Figure \ref{fig:quote_example}). Each quote is matched with the longest sequence of sentences in the post using the Levenshtein edit distance (allowing a distance of 2 characters for typos). A matched text span should contain at least one word and four characters, and cover at least 80\% of the quote to exclude cases where the $>$ symbol is used to quote external content. As a result, 98\% of the matched spans cover the corresponding quotes entirely. Additionally, a sentence in the post is considered to be quoted if at least four non-stopwords appear in a comment's sentence. For example:
\begin{quote}\fontsize{9.4pt}{9.4pt}\selectfont
    Post: \lex{... If you do something, you should be prepared to accept the consequences. ...} \vspace{2pt} \\
    Comment: \lex{... I guess my point is, even if you do believe that ``\textbf{If you do something, you should be prepared to accept the consequences},'' you can still feel bad for the victims. ...}
\end{quote}

We considered manually annotating attacked sentences too, but it turned out to be extremely time-consuming and subjective (Appendix \ref{app:annot_quotes}). We tried to automate it using heuristics (word overlap and vector embeddings), but precision severely deteriorated. As we value the precision of labels over recall, we only use the method described in the previous paragraph. \newcite{Chakrabarty:2019gv} used the same method to collect attack relations in CMV.

\paragraph{Successfully vs. Unsuccessfully Attacked:} After each sentence in a post is labeled as attacked or not, each attacked sentence is further labeled as \emph{successfully attacked} if any of the comments that attack it, or their lower-level comments win a $\Delta$. 

We post-process the resulting labels to increase their validity. First, as a challenger and the OP have discussion down the comment thread, the challenger might attack different sentences than the originally attacked ones and change the OP's view. In this case, it is ambiguous which sentences contribute to the view change. Hence, we extract quotes from all lower-level comments of $\Delta$-winning challengers, and if any of the quotes attack new sentences, this challenger's attacks are excluded from the labeling of \textit{successfully attacked}. This case is not common, however (0.2\%).

Second, if a comment attacks many sentences in the post and change the OP's view, some of them may not contribute to the view change but are still labeled as \textit{successfully attacked}. To reduce this noise, comments that have more than three quotes are excluded from the labeling of \textit{successfully attacked}\footnote{This allows our subsequent analyses to capture stronger signals for successful attacks than without this process.}. This amounts to 12\% of top-level comments (63\% of comments have only one quote, 17\% two quotes, and 8\% three quotes).

Lastly, we verified if quoted sentences are actually attacked. We randomly selected 500 comments and checked if each quoted sentence is purely agreed with without any opposition, challenge, or question. This case was rare (0.4\%)\footnote{Further, this case happened in only one out of the 500 comments (0.2\%), where the author agreed with 4 quoted sentences. In CMV, challengers do use concessions but hardly quote the OP's sentences just to agree.}, so we do not further process this case. Table \ref{tab:data} shows some statistics of the final data.

\begin{table}[t]
    \small
    \centering
    \begin{tabularx}{\linewidth}{XrS[table-format=3.3,input-decimal-markers={,},output-decimal-marker={,}]S[table-format=3.3,input-decimal-markers={,},output-decimal-marker={,}]S[table-format=3.3,input-decimal-markers={,},output-decimal-marker={,}]}\toprule
        Dataset & & {Train} & {Val} & {Test} \\
        \midrule
        \multirow{3}{*}{Attacked}
        & \#posts & 25,839 & 8,763 & 8,558  \\
         & \#sentences & 420,545 & 133,090 & 134,375 \\
         & \#attacked & 119,254 & 40,163 & 40,354 \\
         \midrule
        \multirow{3}{*}{Successful} & \#posts & 3,785 & 1,235  & 1,064 \\ 
          & \#sentences & 66,628 & 20,240 & 17,129 \\
          & \#successful & 8,746 & 2,718 & 2,288 \\
        \bottomrule
    \end{tabularx}
    \caption{Data statistics. ``Attacked'' contains posts with at least one attacked sentence. ``Successful'' contains posts with at least one successfully attacked sentence.}
    \label{tab:data}
\end{table}

\section{Quantifying Sentence Characteristics\label{sec:features}}
As the first step for analyzing the characteristics of attackable sentences, we examine driving reasons for attacks and quantify relevant  characteristics.

\subsection{Rationales and Motivation for Attacks\label{sec:attack_rationales}}

\newcolumntype{x}[1]{>{\centering\arraybackslash}p{#1}}
\begin{table}[t]
    \small
    \centering
    \begin{subtable}[t]{\linewidth}
        \begin{tabularx}{\linewidth}{p{.2cm}X}\toprule
            R1 & \textit{S} is true but does not support the main claim (19\%) \\
            R2 & \textit{S} misses cases suggesting opposite judgment (18\%) \\
            R3 & \textit{S} has exceptions (17\%) \\
            R4 & \textit{S} is false (12\%)  \\
            R5 & \textit{S} misses nuanced distinctions of a concept (8\%) \\
            R6 & \textit{S} is unlikely to happen (6\%) \\
            R7 & \textit{S} has no evidence (6\%) \\
            R8 & \textit{S} uses an invalid assumption or hypothetical (4\%) \\
            R9 & \textit{S} contradicts statements in the argument (4\%) \\
            R10 & Other (4\%) \\
            \bottomrule
        \end{tabularx}
        \caption{Rationales for attacking a sentence (\textit{S}).}
        \label{tab:attack_rationale}
    \end{subtable}\vspace{.3cm}
    \begin{subtable}[t]{\linewidth}
        \begin{tabularx}{\linewidth}{p{.2cm}X}\toprule
            F1 & Personal opinion (28\%) \\ 
            F2 & Invalid hypothetical (26\%) \\
            F3 & Invalid generalization (13\%) \\
            F4 & No evidence (11\%) \\
            F5 & Absolute statement (7\%) \\
            F6 & Concession (5\%) \\
            F7 & Restrictive qualifier (5\%) \\
            F8 & Other (5\%) \\
            \bottomrule
    \end{tabularx}
    \caption{Motivating factors for attacks.}
    \label{tab:attack_motivation}
    \end{subtable}
    \caption{Rationales and motivating factors for attacks.}
\end{table}

To analyze rationales for attacks, two authors examined quotes and rebuttals in the training data (one successful and one unsuccessful comment for each post). From 156 attacks, we identified 10 main rationales (Table \ref{tab:attack_rationale}), which are finer-grained than the refutation reasons in prior work~\cite{Wei:2016ui}.
The most common rationale is that the sentence is factually correct but is irrelevant to the main claim (19\%). Counterexample-related rationales are also common: the sentence misses an example suggesting the opposite judgment to the sentence's own (18\%) and the sentence has exceptions (17\%). 

This analysis is based on polished rebuttals, which mostly emphasize logical aspects, and cannot fully capture other factors that motivate attacks. Hence, we conducted a complementary analysis, where an undergraduate student chose three sentences to attack for each of 50 posts and specified the reasons in their own terms (Table \ref{tab:attack_motivation}). The most common factor is that the sentence is only a personal opinion (28\%). Invalid hypotheticals are also a common factor (26\%). The tone of a sentence motivates attacks as well, such as generalization (13\%), absoluteness (7\%), and concession (5\%).

\subsection{Feature Extraction}
Based on these analyses, we cluster various sentence characteristics into four categories---content, external knowledge, proposition types, and tone.\footnote{Some rationales in Table \ref{tab:attack_rationale} (e.g., R1 and R9) are difficult to operationalize reliably using the current NLP technology and thus are not included in our features.}

\subsubsection{Content}
Content and logic play the most important role in CMV discussions. We extract the content of each sentence at two levels: TFIDF-weighted \textbf{$\boldsymbol{n}$-grams} ($n=1, 2, 3$) and sentence-level \textbf{topics}. Each sentence is assigned one topic using Sentence LDA~\cite{Jo:2011tr}. We train a model on posts in the training set and apply it to all posts, exploring the number of topics $\in \{10, 50, 100\}$.\footnote{We also tried features based on semantic frames using SLING~\cite{Ringgaard:2017vr}, but they were not helpful.}

\subsubsection{External Knowledge\label{sec:knowledge}}
External knowledge sources may provide information as to how truthful or convincing a sentence is (e.g., Table \ref{tab:attack_rationale}-R2, R3, R4, R7 and Table \ref{tab:attack_motivation}-F4). As our knowledge source, we use kialo.com---a collaborative argument platform over more than 1.4K issues. Each issue has a main statement, and users can respond to any existing statement with pro/con statements (1-2 sentences), building an argumentation tree. Kialo has advantages over structured knowledge bases and Wikipedia in that it includes many debatable statements; many attacked sentences are subjective judgments (\S{\ref{sec:attack_rationales}}), so fact-based knowledge sources may have limited utility. In addition, each statement in Kialo has pro/con counts, which may reflect the convincingness of the statement. We scraped 1,417 argumentation trees and 130K statements (written until Oct 2019).

For each sentence in CMV, we retrieve similar statements in Kialo that have at least 5 common words\footnote{Similarity measures based on word embeddings and knowledge representation did not help (Appendix \ref{app:knowledge}).} and compute the following three features. \textbf{Frequency} is the number of retrieved statements; sentences that are not suitable for argumentation are unlikely to appear in Kialo. This feature is computed as $\log_2(N+1)$, where $N$ is the number of retrieved statements. 
\textbf{Attractiveness} is the average number of responses for the matched statements, reflecting how debatable the sentence is. It is computed as $\log_2(M+1)$, where $M = \frac{1}{N} \sum_{i=1}^{N} R_i$ and $R_i$ is the number of responses for the $i$th retrieved statement. 
Lastly, \textbf{extremeness} is $\frac{1}{N} \sum_{i=1}^{N} |P_i - N_i|$, where $P_i$ and $N_i$ are the proportions (between 0 and 1) of pro responses and con responses for the $i$th retrieved statement. A sentence that most people would see flawed would have a high extremeness value.

\subsubsection{Proposition Types}
Sentences convey different types of propositions, such as predictions and hypotheticals. No proposition types are fallacious by nature, but some of them may make it harder to generate a sound argument. They also communicate different moods, causing the hearer to react differently. We extract 13 binary features for proposition types. They are all based on lexicons and regular expressions, which are available in Appendix \ref{app:lexicons}.

\textbf{Questions} express the intent of information seeking. Depending on the form, we define three features: \textbf{confusion} (e.g., \lex{I don't understand}), \textbf{why/how} (e.g., \lex{why ...?}), and \textbf{other}. 

\textbf{Normative} sentences suggest that an action be carried out. Due to their imperative mood, they can sound face-threatening and thus attract attacks. 

\textbf{Prediction} sentences predict a future event. They can be attacked with reasons why the prediction is unlikely (Table \ref{tab:attack_rationale}-R6), as in critical questions for \textit{argument from cause to effect}~\cite{Walton:2008schem}.

\textbf{Hypothetical} sentences may make implausible assumptions (Table \ref{tab:attack_rationale}-R8 and Table \ref{tab:attack_motivation}-F2) or restrict the applicability of the argument too much (Table \ref{tab:attack_motivation}-F7). 

\textbf{Citation} often strengthens a claim using authority, but the credibility of the source could be attacked~\cite{Walton:2008schem}. 

\textbf{Comparison} may reflect personal preferences that are vulnerable to attacks (Table \ref{tab:attack_motivation}-F1). 

\textbf{Examples} in a sentence may be attacked for their invalidity~\cite{Walton:2008schem} or counterexamples (Table \ref{tab:attack_rationale}-R3). 

\textbf{Definitions} form a ground for arguments, and challengers could undermine an argument by attacking this basis (e.g., Table \ref{tab:attack_rationale}-R5).

\textbf{Personal stories} are the arguer's experiences, whose validity is difficult to refute. A sentence with a personal story has subject \lex{I} and a non-epistemic verb; or it has \lex{my} modifying non-epistemic nouns.

\textbf{Inclusive sentences} that mention \textbf{\lex{you}} and \textbf{\lex{we}} engage the hearer into the discourse~\cite{Hyland:2005wh}, making the argument more vulnerable to attacks.

\subsubsection{Tone}
Challengers are influenced by the tone of an argument, e.g., subjectiveness, absoluteness, or confidence (Table \ref{tab:attack_motivation}). We extract 8 features for the tone of sentences. 

\textbf{Subjectivity} comprises judgments, which are often attacked due to counterexamples (Table \ref{tab:attack_rationale}-R2) or their arbitrariness (Table \ref{tab:attack_motivation}-F1, \citet{Walton:2008schem}). The subjectivity of a sentence is the average subjectivity score of words based on the Subjectivity Lexicon~\cite{Wilson:2005subjlex} (non-neutral words of ``weaksubj'' = 0.5 and ``strongsubj'' = 1).

\textbf{Concreteness} is the inverse of abstract diction, whose meaning depends on subjective perceptions and experiences. The concreteness of a sentence is the sum of the standardized word scores based on \citet{Brysbaert:2014eo}'s concreteness lexicon.

\textbf{Qualification} expresses the level of generality of a claim, where absolute statements can motivate attacks (Table \ref{tab:attack_motivation}-R3). The qualification score of a sentence is the average word score based on our lexicon of qualifiers and generality words. 

\textbf{Hedging} can sound unconvincing~\cite{Durik:2008kd} and motivate attacks. A sentence's hedging score is the sum of word scores based on our lexicon of downtoners and boosters.

\textbf{Sentiment} represents the valence of a sentence. Polar judgments may attract more attacks than neutral statements. We calculate the sentiment of each sentence with BERT~\cite{Devlin:2018bert} trained on the data of SemEval 2017 Task 4~\cite{semeval17task4}. \textbf{Sentiment score} is a continuous value ranging between -1 (negative) and +1 (positive), and \textbf{sentiment categories} are nominal (positive, neutral, and negative)\footnote{We achieved an average recall of 0.705, which is higher than the winner team's performance of 0.681.}.
In addition, we compute the scores of \textbf{arousal} (intensity) and \textbf{dominance} (control) as the sum of the standardized word scores based on \citet{Warriner:2013gv}'s lexicon.

\section{Task 1: Attackability Characteristics\label{sec:task1}}
One of our goals in this paper is to analyze what characteristics of sentences are associated with a sentence's attackability. 
Hence, in this section, we measure the effect size and statistical significance of each feature toward two labels: (i) whether a sentence is attacked or not, using the dev set of the ``Attacked'' dataset ($N$=553,635), (ii) whether a sentence is attacked successfully or unsuccessfully, using all attacked sentences ($N$=159,417).\footnote{Simply measuring the predictive power of features in a prediction setting provides an incomplete picture of the roles of the characteristics. Some features may not have drastic contribution to prediction due to their infrequency, although they may have significant effects on attackability.}
Since the effects of characteristics may depend on the issue being discussed, the effect of each feature is estimated conditioned on the domain of each post using a logistic regression, and the statistical significance of the effect is assessed using the Wald test.
For interpretation purposes, we use \textit{odds ratio} (OR)---the exponent of the effect size.\footnote{Odds are the ratio of the probability of a sentence being (successfully) attacked to the probability of being not (successfully) attacked; OR is the ratio of odds when the value of the characteristic increases by one unit (Appendix \ref{app:stat_models}).}

\newcommand{\bi}{$^\dagger$}
\newcommand{\scale}{$^\ddagger$}
\begin{table}[t]
    \small
    \centering
    \begin{tabularx}{\linewidth}{p{.4mm}p{3.35cm}cc} \toprule
         & Feature & Attacked & Successful \\ \midrule
        \rotate{5}{Content} & Topic47: Gender\bi{} & \plus{ 1.37 } \threestars{} & \plus{ 1.34 } \threestars{} \\
         & Topic8: Race\bi{} & \plus{ 1.19 } \threestars{} & \plus{ 1.21 } \twostars{} \\
         & Topic6: Food\bi{} & \plus{ 1.00 } \nostar{} & \plus{ 1.39 } \threestars{} \\
         & Topic38: Movie \& Show\bi{} & \plus{ 1.03 } \nostar{} & \minus{ 0.78 } \threestars{} \\
         & Topic4: CMV-Specific\bi{} & \minus{ 0.16 } \threestars{} & \minus{ 0.36 } \twostars{} \\
        \midrule
        \rotate{3}{\scriptsize Knowledge} & Kialo Frequency (log2) & \plus{ 1.18 } \threestars{} & \plus{ 1.07 } \threestars{} \\
         & Kialo Attractiveness (log2) & \plus{ 1.30 } \threestars{} & \plus{ 1.18 } \threestars{} \\
         & Kialo Extremeness & \plus{ 1.51 } \threestars{} & \plus{ 1.19 } \threestars{} \\
        \midrule
        \rotate{13}{Proposition Types} & Question - Confusion\bi{} & \minus{ 0.97 } \nostar{} & \plus{ 1.29 } \onestar{} \\
         & Question - Why/How\bi{} & \plus{ 1.77 } \threestars{} & \plus{ 1.27 } \threestars{} \\
         & Question - Other\bi{} & \plus{ 1.16 } \threestars{} & \plus{ 1.11 } \onestar{} \\
         & Citation\bi{} & \minus{ 0.53 } \threestars{} & \plus{ 1.17 } \onestar{} \\
         & Definition\bi{} & \plus{ 1.04 } \nostar{} & \plus{ 1.32 } \twostars{} \\
         & Normative\bi{} & \plus{ 1.26 } \threestars{} & \plus{ 1.10 } \twostars{} \\
         & Prediction\bi{} & \plus{ 1.22 } \threestars{} & \plus{ 1.02 } \nostar{} \\
         & Hypothetical\bi{} & \plus{ 1.29 } \threestars{} & \plus{ 1.07 } \nostar{} \\
         & Comparison\bi{} & \plus{ 1.25 } \threestars{} & \plus{ 1.02 } \nostar{} \\
         & Example\bi{} & \plus{ 1.20 } \threestars{} & \plus{ 1.17 } \onestar{} \\
         & Personal Story\bi{} & \minus{ 0.70 } \threestars{} & \plus{ 1.09 } \twostars{} \\
         & Use of \textit{You}\bi{} & \plus{ 1.18 } \threestars{} & \plus{ 1.04 } \nostar{} \\
         & Use of \textit{We}\bi{} & \plus{ 1.24 } \threestars{} & \minus{ 0.98 } \nostar{} \\
        \midrule
        \rotate{10}{Tone} & Subjectivity\scale{} & \plus{ 1.03 } \threestars{} & \minus{ 0.97 } \threestars{} \\
         & Concreteness\scale{} & \minus{ 0.87 } \threestars{} & \minus{ 0.92 } \threestars{} \\
         & Hedges\scale{} & \plus{ 1.04 } \threestars{} & \plus{ 1.06 } \threestars{} \\
         & Quantification\scale{} & \minus{ 0.97 } \threestars{} & \plus{ 1.02 } \nostar{} \\
         & Sentiment Score\scale{} & \minus{ 0.87 } \threestars{} & \minus{ 1.00 } \nostar{} \\
         & Sentiment: Positive\bi{} & \minus{ 0.76 } \threestars{} & \minus{ 0.99 } \nostar{} \\
         & Sentiment: Neutral\bi{} & \minus{ 0.82 } \threestars{} & \minus{ 1.00 } \nostar{} \\
         & Sentiment: Negative\bi{} & \plus{ 1.34 } \threestars{} & \plus{ 1.00 } \nostar{} \\
         & Arousal\scale{} & \plus{ 1.02 } \threestars{} & \minus{ 0.95 } \threestars{} \\
         & Dominance\scale{} & \plus{ 1.07 } \threestars{} & \plus{ 1.08 } \threestars{} \\
        \bottomrule 
    \end{tabularx}
    \caption{Odds ratio (OR) and statistical significance of features. An effect is positive (\plus{blue}) if OR $>$ 1 and negative (\minus{red}) if OR $<$ 1. (\bi{}: binary, \scale{}: standardized / *: $p<0.05$, **: $p<0.01$, ***: $p<0.001$)}
    \label{tab:feature_effects}
\end{table}

\subsection{Content} 
Attacked sentences tend to mention big issues like gender, race, and health as revealed in topics 47, 8, and 6 (Table \ref{tab:feature_effects}) and $n$-grams \lex{life}, \lex{weapons}, \lex{women}, \lex{society}, and \lex{men} (Table \ref{tab:ngrams_effects_full} in Appendix \ref{app:ngrams}). These issues are also positively correlated with successful attacks. On the other hand, mentioning relatively personal issues (\lex{tv}, \lex{friends}, topic 38) seems negatively correlated with successful attacks. So do forum-specific messages (\lex{cmv}, \lex{thank}, topic 4).

Attacking seemingly evidenced sentences appears to be effective for persuasion when properly done. Successfully attacked sentences are likely to mention specific data (\lex{data}, \lex{\%}) and be the OP's specific reasons under bullet points (\lex{2.} and \lex{3.}). 

$n$-grams capture various characteristics that are vulnerable to attacks, such as uncertainty and absoluteness (\lex{i believe}, \lex{never}), hypotheticals (\lex{if i}), questions (\lex{?}, \lex{why}), and norms (\lex{should}). 


\subsection{External Knowledge} 
The Kialo-based knowledge features provide significant information about whether a sentence would be attacked successfully (Table \ref{tab:feature_effects}). As the frequency of matched statements in Kialo increases twice, the odds for successful attack increase by 7\%. 
As an example, the following attacked sentence has 18 matched statements in Kialo.
\begin{quote}\fontsize{9.4pt}{9.4pt}\selectfont
    \lex{I feel like it is a parents right and responsibility to make important decisions for their child.}
\end{quote}

The attractiveness feature has a stronger effect; as matched statements have twice more responses, the odds for successful attack increase by 18\%, probably due to higher debatability.

A sentence being completely extreme (i.e., the matched sentences have only pro or con responses) increases the odds for successful attack by 19\%. 

As expected, the argumentative nature of Kialo allows its statements to match many subjective sentences in CMV and serves as an effective information source for a sentence's attackability.

\subsection{Proposition Types} 
Questions, especially why/how, are effective targets for successful attack (Table \ref{tab:feature_effects}). Although challengers do not pay special attention to expressions of confusion (see column ``Attacked''), they are positively correlated with successful attack (OR=1.29). 

Citations are often used to back up an argument and have a low chance of being attacked, reducing the odds by half. However, properly attacking citations significantly increases the odds for successful attack by 17\%. Similarly, personal stories have a low chance of being attacked and definitions do not attract challengers' attacks, but attacking them is found to be effective for successful persuasion.

All other features for proposition types have significantly positive effects on being attacked (OR=1.18--1.29), but only normative and example sentences are correlated with successful attack.

\subsection{Tone} 
Successfully attacked sentences tend to have lower subjectivity and arousal (Table \ref{tab:feature_effects}), in line with the previous observation that they are more data- and reference-based than unsuccessfully attacked sentences. In contrast, sentences about concrete concepts are found to be less attackable. 

Uncertainty (high hedging) and absoluteness (low qualification) both increase the chance of attacks, which aligns with the motivating factors for attacks (Table \ref{tab:attack_motivation}), while only hedges are positively correlated with successful attacks, implying the importance of addressing the arguer's uncertainty.

Negative sentences with high arousal and dominance have a high chance of being attacked, but most of these characteristics have either no or negative effects on successful attacks.

\subsection{Discussion}
We have found some evidence that, somewhat counter-intuitively, seemingly evidenced sentences are more effective to attack. Such sentences use specific data (\lex{data}, \lex{\%}), citations, and definitions. Although attacking these sentences may require even stronger evidence and deeper knowledge, arguers seem to change their viewpoints when a fact they believe with evidence is undermined. In addition, it seems very important and effective to identify and address what the arguer is confused (confusion) or uncertain (hedges) about.

Our analysis also reveals some discrepancies between the characteristics of sentences that challengers commonly think are attackable and those that are indeed attackable. Challengers are often attracted to subjective and negative sentences with high arousal, but successfully attacked sentences have rather lower subjectivity and arousal, and have no difference in negativity compared to unsuccessfully attacked sentences. Furthermore, challengers pay less attention to personal stories, while successful attacks address personal stories more often.

\section{Task 2: Attackability Prediction}
Now we examine how well computational models can detect attackable sentences in arguments.

\subsection{Problem Formulation}
This task is cast as ranking sentences in each post by their attackability scores predicted by a regression model. We consider two types of attackability: (i) whether a sentence will be attacked or not, (ii) whether a sentence will be successfully attacked or not (attacked unsuccessfully + unattacked). For both settings, we consider posts that have at least one sentence with the positive label (Table \ref{tab:data}).

We use three evaluation metrics. \textbf{P@1} is the precision of the first ranked sentence, measuring the model's accuracy when choosing one sentence to attack for each post. Less strictly, \textbf{A@3} gives a score of 1 if any of the top 3 sentences is a positive instance and 0 otherwise. \textbf{AUC} measures individual sentence-level accuracy---how likely positive sentences are assigned higher probabilities. 

\subsection{Comparison Models}

For machine learning models, we explore two logistic regression models to compute the probability of the positive label for each sentence, which becomes the sentence's \textit{attackability score}. 
\textbf{LR} is a basic logistic regression with our features\footnote{We tried the number of topics $\in$ \{10, 50, 100\}, and 50 has the best AUC on the val set for both prediction settings.} (Section \ref{sec:features}) and binary variables for domains. We explored feature selection using L1-norm and regularization using L2-norm.\footnote{We also tried a multilayer perceptron to model feature interactions, but it consistently performed worse than LR.} 
\textbf{BERT} is logistic regression where our features are replaced with the BERT embedding of the input sentence~\cite{Devlin:2018bert}. Contextualized BERT embeddings have achieved state-of-the-art performance in many NLP tasks. 
We use the pretrained, uncased base model from Hugging Face~\cite{huggingface2019} and fine-tune it during training.\footnote{Details for reproducibility are in Appendix \ref{sec:reproducibility}.}

We explore two baseline models. \textbf{Random} is to rank sentences randomly. \textbf{Length} is to rank sentences from longest to shortest, with the intuition that longer sentences may contain more information and thus more content to attack as well.

Lastly, we estimate laypeople's performance on this task. Three undergraduate students each read 100 posts and rank three sentences to attack for each post. Posts that have at least one positive instance are randomly selected from the test set.\footnote{We were interested in the performance of young adults who are academically active and have a moderate level of life experience. Their performance may not represent the general population, though.}

\subsection{Results\label{sec:pred_results}}
\begin{table}[t]
    \small
    \centering
    \begin{tabularx}{\linewidth}{lCCCCCC} \toprule
         & \multicolumn{3}{c}{Attacked} & \multicolumn{3}{c}{Successful} \\ \cmidrule{2-4} \cmidrule(l){5-7}
         & P@1 & A@3 & AUC & P@1 & A@3 & AUC \\ 
        \cmidrule{1-4} \cmidrule(l){5-7}
        Random & 35.9 & 66.0 & 50.1 & 18.9 & 45.0 & 50.1 \\
        Length & 42.9 & 73.7 & 54.5 & 22.3 & 52.1 & 55.7 \\
        \cmidrule{1-4} \cmidrule(l){5-7}
        LR & 47.1 & 76.2 & 61.7 & 24.2 & 54.5 & 59.3 \\
        ~($\times$) Content & 45.2 & 74.4 & 58.1 & 24.0 & 52.6 & 57.0 \\
        ~($\times$) Knowledge & 47.0 & 76.0 & 61.7 & 24.1 & 54.3 & 59.0 \\
        ~($\times$) Prop Type & 46.7 & 75.9 & 61.5 & 24.4 & 53.6 & 59.0 \\
        ~($\times$) Tone & 47.0 & 76.0 & 61.9 & 25.2 & 56.2 & 59.4 \\
        BERT & 49.6 & 77.8 & 64.4 & 28.3 & 57.2 & 62.0 \\
        \cmidrule{1-4} \cmidrule(l){5-7}
        Humans$^\dagger$ & 51.7 & 80.1 & -- & 27.8 & 54.2 & --  \\
        \bottomrule
    \end{tabularx}
    \caption{Prediction accuracy. All LR/BERT scores (rows 3--8) have standard deviations between 0.1 and 1.1, significantly outperforming ``Length''. $^\dagger$The average bootstrap accuracy after resampling 100K times with sample size 200
    ---the standard deviations of P@1 and A@3 range between 2.1 and 3.5.}
    \label{tab:pred_acc}
\end{table}

All computational models were run 10 times, and their average accuracy is reported in Table \ref{tab:pred_acc}. Both the LR and BERT models significantly outperform the baselines, while the BERT model performs best. For predicting attacked sentences, 
the BERT model's top 1 decisions match the gold standard 50\% of the time; its decisions match 78\% of the time when three sentences are chosen.
Predicting successfully attacked sentences is harder, but the performance gap between our models and the baselines gets larger. The BERT model's top 1 decisions match the gold standard 28\% of the time---a 27\% and 10\% boost from random and length-based performance, respectively.

To examine the contribution of each feature category, we did ablation tests based on the best performing LR model (Table \ref{tab:pred_acc} rows 4--7). The two prediction settings show similar tendencies. Regarding P@1 for successful attack, content has the highest contribution, followed by knowledge, proposition types, and tone. This result reaffirms the importance of content for a sentence's attackability. But the other features still have significant contribution, yielding higher P@1 and AUC (Table \ref{tab:pred_acc} row 4) than the baselines.

It is worth noting that our features, despite the lower accuracy than the BERT model, are clearly informative of attackability prediction as Table \ref{tab:pred_acc} row 3 shows. Moreover, since they directly operationalize the sentence characteristics we compiled, it is pretty transparent that they capture relevant information that contributes to sentence attackability and help us better understand what characteristics have positive and negative signals for sentence attackability. We speculate that transformer models like BERT are capable of encoding these characteristics more sophisticatedly and may include some additional information, e.g., lexical patterns, leading to higher accuracy. But at the same time, it is less clear exactly what they capture and whether they capture relevant information or irrelevant statistics, as is often the case in computational argumentation \cite{Niven:2019cq}.

\begin{figure}[t]
    \centering
    \includegraphics[width=\linewidth]{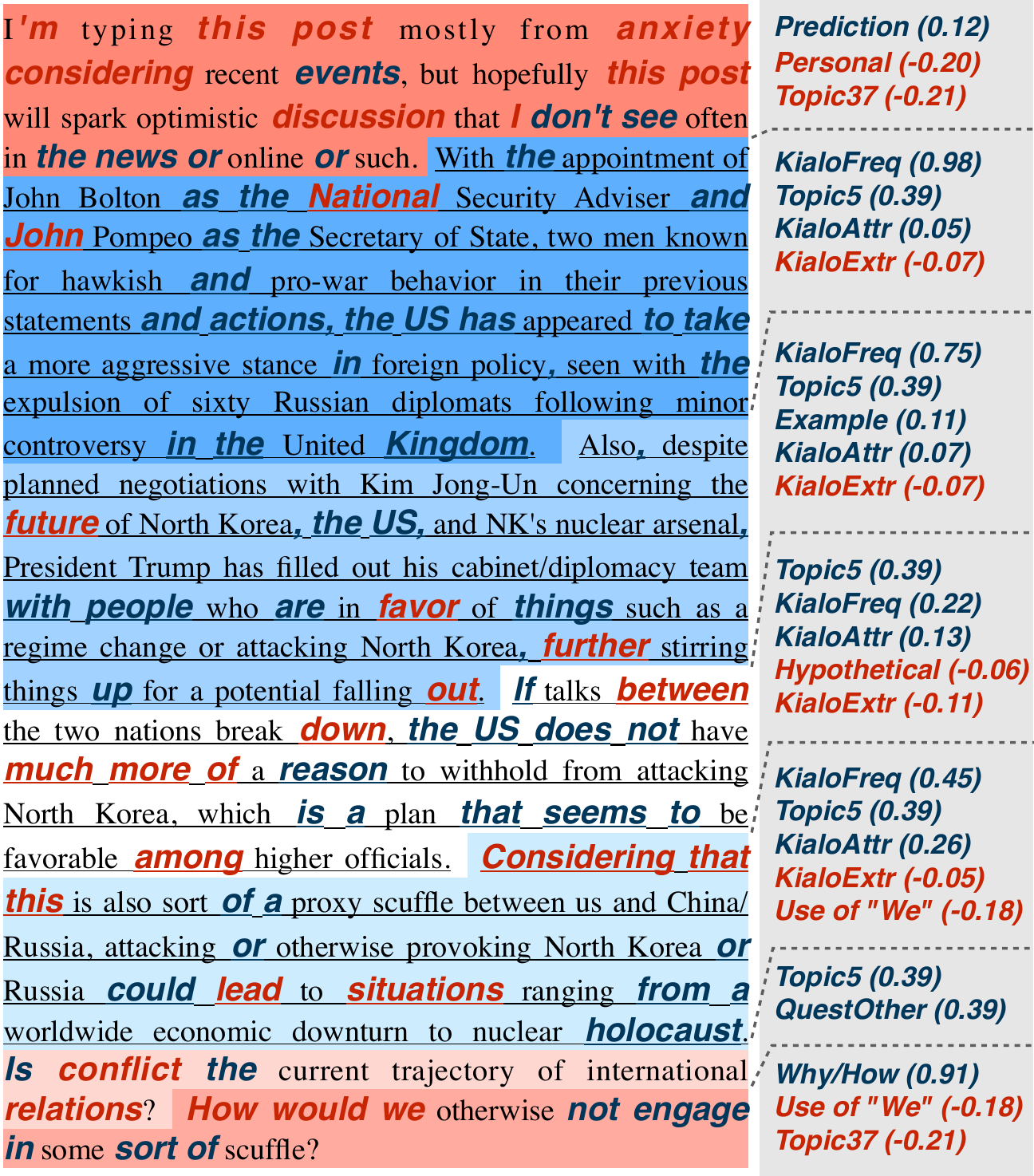} 
    \caption{Prediction visualization. Background color indicates predicted attackability (\textcolor{royalblue}{blue}: high, \textcolor{orangered}{red}: low). Successfully attacked sentences are underlined. Features with high/low weights are indicated with \plus{blue}/\minus{red}.}
    \label{fig:pred_example}
\end{figure}

Figure \ref{fig:pred_example} illustrates how LR allows us to interpret the contribution of different features to attackability, by visualizing a post with important features highlighted.
For instance, external knowledge plays a crucial role in this post; all successfully attacked sentences match substantially more Kialo statements than other sentences. The attackability scores of these sentences are also increased by the use of hypotheticals and certain $n$-grams like \lex{could}.
These features align well with the actual attacks by successful challengers. For instance, they pointed out that the expulsion of Russian diplomats (sentence 2) is not an aggressive reaction because the diplomats can be simply replaced with new ones. Kialo has a discussion on the relationship between the U.S. and Russia, and one statement puts forward exactly the same point that the expulsion was a forceful-looking but indeed a nice gesture. Similarly, a successful challenger pointed out the consistent attitude of the U.S. toward regime change in North Korea (sentence 3), and the North Korean regime is a controversial topic in Kialo. Lastly, one successful challenger attacked the hypothetical outcomes in sentences 4 and 5, pointing out that those outcomes are not plausible, and the LR model also captures the use of hypothetical and the word \lex{could} as highly indicative of attackability.
More successful and erroneous cases are in Appendix \ref{app:visualization_examples}.

Laypeople perform significantly better than the BERT model for predicting attacked sentences, but only comparably well for successfully attacked sentences (Table \ref{tab:pred_acc} row 9).
Persuasive argumentation in CMV requires substantial domain knowledge, but laypeople do not have such expertise for many domains. The BERT model, however, seems to take advantage of the large data and encodes useful linguistic patterns that are predictive of attackability.
A similar tendency has been observed in predicting persuasive refutation~\cite{Guo.2020.WWW}, where a machine-learned model outperformed laypeople.
Nevertheless, in our task, the humans and the BERT model seem to make similar decisions; the association between their choices of sentences is high, with odds ratios ranging between 3.43 (top 1) and 3.33 (top 3). Interestingly, the LR model has a low association with the human decisions for top 1 (OR=2.65), but the association exceeds the BERT model for top 3 (OR=3.69). It would be interesting to further examine the similarities and differences in how humans and machines choose sentences to attack.



\section{Conclusion}
We studied how to detect attackable sentences in arguments for successful persuasion. Using online arguments, we demonstrated that a sentence's attackability is associated with many of its characteristics regarding its content, proposition types, and tone, and that Kialo provides useful information about attackability. Based on these findings we demonstrated that machine learning models can automatically detect attackable sentences, comparably well to laypeople.

Our work contributes a new application to the growing literature on causal inference from text \cite{egami2018make}, in the setting of ``text as a treatment''. Specifically, our findings in Section \ref{sec:task1} pave the way towards answering the causal question: would attacking a certain type of sentence (e.g., questions or expressions of confusion) in an argument increase the probability of persuading the opinion holder?
While our findings suggest initial hypotheses about the characteristics of sentences that can be successfully attacked, establishing causality in a credible manner would require addressing confounders, such as the challenger's reputation~\cite{manzoor2020ethos} and persuasive skill reflected in their attack~\cite{Tan:2014ut}. We leave this analysis to future work.



Our work could be improved also by including discourse properties (coherence, cohesiveness). Further, argumentation structure (support relations between sentences or lack thereof) might provide useful information about each sentence's attackability.

\section*{Acknowledgments}
This research was supported by the Kwanjeong Educational Foundation.


\bibliography{emnlp2020}
\bibliographystyle{acl_natbib}

\newpage

\appendix

\onecolumn
\section{Annotating Attacked Sentences\label{app:annot_quotes}}
We tried capturing sentences in posts that are addressed by comments but not directly quoted. To see its feasibility, we randomly sampled 100 post-comment pairs that do not contain direct quotes and then asked an undergraduate native speaker of English (who has no knowledge about this work) to mark attacked sentences in each post, if any. This revealed two challenges. First, human annotation is subjective when compared to a co-author's result and very time-consuming (2.5 min/comment). Second, we tried several methods to automatically identify attacked sentences. We compared the similarity between each post sentence with the comment (first sentence of the comment, first sentence of each paragraph, or all comment text) based on word overlap with/without synonym expansion and the GloVe embeddings. But it turned out to be difficult to get similar results to human annotations. Therefore, we decided to use only those sentences that are direct quoted or have at least 4 common words with a comment's sentence as the most reliable labels.

\newpage

\section{External Knowledge\label{app:knowledge}}
In this section, we describe the methods that we explored to use Kialo as a knowledge base but that were not successful.

\subsection{UKP Sentence Embedding-Based Retrieval}
We measured the similarity between CMV sentences and Kialo statements using the UKP sentence embedding---BERT embeddings fine-tuned to measure argument similarity \cite{reimers-etal-2019-classification}. Specifically, the authors provide pretrained embeddings constructed by appending a final softmax layer to BERT to predict a numerical \textit{dissimilarity} score between 0 and 1 for each sentence pair in the UKP ASPECT corpus. The 3,595 sentence pairs in this corpus were drawn from 28 controversial topics and annotated via crowd workers to be ``unrelated'' or of ``no'', ``some'' or ``high'' similarity. They report a mean F1-score of 65.39\% on a held-out subset of this corpus, which was closest to human performance (F1=78.34\%) among all competing methods that were not provided with additional information about the argument topic.

We used this fine-tuned model to measure the dissimilarity between each CMV sentence and Kialo statements. Based on this information, we extracted the feature \textbf{UKP Avg Distance 10}, which is the average dissimilarity score of the 10 Kialo statements that are closest to the sentence. This score is expected to be low if a sentence has many similar statements in Kialo. In addition, we extracted the same \textbf{frequency}, \textbf{attractiveness}, and \textbf{extremeness} features as in \S{\ref{sec:knowledge}}. Here, we determine whether a CMV sentence and a Kialo statement are ``matched'' based on several dissimilarity thresholds (0.1, 0.2, 0.3, 0.4); A Kialo statement is considered matched with a CMV sentence if the dissimilarity is below the selected threshold.

\subsection{Semantic Frame-Based Knowledge}
We extracted semantic frames from CMV sentences and Kialo statements, using Google SLING~\cite{Ringgaard:2017vr}. For each frame in a sentence or statement, a ``knowledge piece'' is defined as the concatenation of the predicate and arguments (except negation); the predicate is lemmatized and the arguments are stemmed to remove differences in verb/noun forms. We also mark each knowledge piece as negated if the frame contains negation. Example knowledge pieces include:
\begin{itemize}[topsep=0pt,itemsep=-4pt]
    \item ARG0:peopl-ARG1:right-ARGM-MOD:should-PRED:have (Negation: true)
    \item ARG1:person-ARG2:abl-ARGM-MOD:should-PRED:be (Negation: false)
\end{itemize}

For each CMV sentence, we extracted two features: the count of knowledge pieces in Kialo that are \textbf{consistent} with those in the sentence, and the count of knowledge pieces in Kialo that are \textbf{conflicting} with those in the sentence. Two knowledge pieces are considered consistent if they are identical, and conflicting if they are identical but negated. Attackable sentences are expected to have many consistent and conflicting knowledge pieces in Kialo. If we assume that most statements in Kialo are truthful, attackable sentences may have more conflicting knowledge pieces than consistent knowledge pieces.

\subsection{Word Sequence-Based Knowledge}
Treating each frame as a separate knowledge piece does not capture the dependencies between multiple predicates within a sentence. Hence, we tried a simple method to capture this information, where a knowledge pieces is defined as the concatenation of verbs, nouns, adjectives, modal, prepositions, subordinating conjunctions, numbers, and existential \lex{there} within a sentence; but independent clauses (e.g., a \lex{because} clause) were separated off. All words were lemmatized. Each knowledge piece is negated if the source text has negation words. Example knowledge pieces include:
\begin{itemize}[topsep=0pt,itemsep=-4pt]
    \item gender-be-social-construct (Negation: true)
    \item congress-shall-make-law-respect-establishment-of-religion-prohibit-free-exercise (Negation: false)
\end{itemize}

For each CMV sentence, we extracted the same two features as in semantic frame-based knowledge pieces: the count of knowledge pieces in Kialo that are \textbf{consistent} with those in the sentence, and the count of knowledge pieces in Kialo that are \textbf{conflicting} with those in the sentence.

\subsection{Effects and Statistical Significance}
The effects and statistical significance of the above features were estimated in the same way as \S{\ref{sec:task1}} and are shown in Table \ref{tab:kialo_effects}. Word sequence-based knowledge has no effect, probably because not many knowledge pieces are matched. Most of the other features have significant effects only for ``Attacked''. We speculate that a difficulty comes from the fact that both vector embedding-based matching and frame-based matching are inaccurate in many cases. UKP sentence embeddings often retrieve Kialo statements that are only topically related to a CMV sentence. Similarly, frame-based knowledge pieces often cannot capture complex information conveyed in a CMV sentence. In contrast, word overlap-based matching seems to be more reliable and better retrieve Kialo statements that have similar content to a CMV sentence.

\begin{table}[!htbp]
    \small
    \centering
    \begin{tabularx}{.6\linewidth}{p{5cm}cc} \toprule
        Knowledge Feature & Attacked & Successful \\ 
        \midrule
        Word Overlap Frequency (log2) & \plus{ 1.18 } \threestars{} & \plus{ 1.07 } \threestars{} \\
        Word Overlap Attractiveness (log2) & \plus{ 1.30 } \threestars{} & \plus{ 1.18 } \threestars{} \\
        Word Overlap Extremeness & \plus{ 1.51 } \threestars{} & \plus{ 1.19 } \threestars{} \\
        \midrule
        UKP Avg Distance 10\scale{} & \minus{ 0.93 } \threestars{} & \minus{ 0.98 } \onestar{} \\
        UKP 0.1 Frequency\bi{} & \plus{ 1.08 } \onestar{} & \minus{ 0.99 } \nostar{} \\
        UKP 0.1 Attractiveness\bi{} & \plus{ 1.11 } \onestar{} & \plus{ 1.08 } \nostar{} \\
        UKP 0.1 Extremeness & \plus{ 3.49 } \onestar{} & \plus{ 6.77 } \nostar{} \\
        UKP 0.2 Frequency\bi{} & \plus{ 1.02 } \twostars{} & \plus{ 1.01 } \nostar{} \\
        UKP 0.2 Attractiveness\bi{} & \plus{ 1.05 } \threestars{} & \plus{ 1.06 } \nostar{} \\
        UKP 0.2 Extremeness & \plus{ 1.69 } \threestars{} & \plus{ 1.76 } \nostar{} \\
        UKP 0.3 Frequency\bi{} & \plus{ 1.04 } \threestars{} & \plus{ 1.01 } \nostar{} \\
        UKP 0.3 Attractiveness\bi{} & \plus{ 1.09 } \threestars{} & \plus{ 1.02 } \nostar{} \\
        UKP 0.3 Extremeness & \plus{ 2.44 } \threestars{} & \plus{ 1.40 } \nostar{} \\
        UKP 0.4 Frequency\bi{} & \plus{ 1.04 } \threestars{} & \plus{ 1.01 } \twostars{} \\
        UKP 0.4 Attractiveness\bi{} & \plus{ 1.12 } \threestars{} & \plus{ 1.01 } \nostar{} \\
        UKP 0.4 Extremeness & \plus{ 2.35 } \threestars{} & \plus{ 1.02 } \nostar{} \\
        \midrule
        Frame Knowledge Consistent & \plus{ 1.28 } \threestars{} & \plus{ 1.01 } \nostar{} \\
        Frame Knowledge Conflict & \plus{ 1.37 } \threestars{} & \plus{ 1.08 } \nostar{} \\
        \midrule
        Word Sequence Knowledge Consistent & \plus{ 1.05 } \nostar{} & \minus{ 0.98 } \nostar{} \\
        Word Sequence Knowledge Conflict & \plus{ 1.18 } \nostar{} & \plus{ 1.49 } \nostar{} \\
        \bottomrule
    \end{tabularx}
    \caption{Odds ratio (OR) and statistical significance of features. An effect is positive (\plus{blue}) if OR $>$ 1 and negative (\minus{red}) if OR $<$ 1. (\bi{}: log2, \scale{}: standardized / *: $p<0.05$, **: $p<0.01$, ***: $p<0.001$)}
    \label{tab:kialo_effects}
\end{table}

\newpage
\section{Lexicons\label{app:lexicons}}
Table \ref{tab:feature_lexicons} shows the lexicons and regular expressions used in feature extraction. \regex{pattern} represents a regular expression.

\begin{table}[!hp]
    \small
    \centering
    \begin{tabularx}{\linewidth}{p{3.6cm}X} \toprule
        Feature & Pattern \\ 
        \midrule
        Question - Confusion & \regex{\verb=(^| )i (\S+ ){,2}(not|n't|never) (understand|know)=}, \regex{\verb=(not|n't) make sense=}, \regex{\verb=(^| )i (\S+ ){,2}(curious|confused)=}, \regex{\verb=(^| )i (\S+ ){,2}wonder=}, \regex{\verb=(me|myself) wonder=} \\
        Question - Why/How & \regex{\verb=(^| )(why|how).*\?=} \\
        Question - Other & ? \\
        Normative &  should, must, ``(have$|$has) to'', ``have got to'', ``'ve got to'', gotta, need, needs \\
        Prediction & \regex{\verb=(am$|$'m$|$are$|$'re$|$is$|$'s) (not )?(going to$|$gonna)=}, will, won't, would, shall \\
        Hypothetical & \regex{\verb=(^|, )if|unless=} \\
        Citation & \regex{\verb= {PATTERN} that [^.,!?]=} (\verb=PATTERN=:
        said, reported, mentioned, declared, claimed, admitted, explained, insisted, promised, suggested, recommended, denied, blamed, apologized, agreed, answered, argued, complained, confirmed, proposed, replied, stated, told, warned, revealed), according to, \regex{\verb=https?:=} \\
        Comparison & than, compared to \\
        Examples & \regex{\verb=(^| )(for example|for instance|such as|e\.g\.)( |$)=} \\
        Definition & define, definition \\
        Personal Story & \textbf{Epistemic verbs:} think, believe, see, know, feel, say, understand, mean, sure, agree, argue, consider, guess, realize, hope, support, aware, disagree, post, mention, admit, accept, assume, convince, wish, appreciate, speak, suppose, doubt, explain, wonder, discuss, view, suggest, recognize, respond, acknowledge, clarify, state, sorry, advocate, propose, define, apologize, curious, figure, claim, concede, debate, list, oppose, describe, suspect, reply, bet, realise, defend, convinced, offend, concern, intend, certain, conclude, reject, challenge, thank, condone, value, skeptical, contend, anticipate, maintain, justify, recommend, confident, promise, guarantee, comment, unsure, elaborate, posit, swear, dispute, imply, misunderstand. \textbf{Epistemic nouns:} view, opinion, mind, point, argument, belief, post, head, position, reasoning, understanding, thought, reason, question, knowledge, perspective, idea, way, stance, vote, best, cmv, response, definition, viewpoint, example, claim, logic, conclusion, thinking, comment, statement, theory, bias, assumption, answer, perception, intention, contention, word, proposal, thesis, interpretation, reply, guess, evidence, explanation, hypothesis, assertion, objection, criticism, worldview, impression, apology, philosophy  \\
        Use of \lex{You} & you, your, yours \\
        Use of \lex{We} & \regex{\verb=(^| )we |(?<!the) (us|our|ours)( |$)=} \\
        \midrule
        Subjectivity & \newcite{Wilson:2005subjlex} \\
        Concreteness & \newcite{Brysbaert:2014eo} \\
        Hedges & \textbf{Downtoners (score=1):} allegedly, apparently, appear to, conceivably, could be, doubtful, fairly, hopefully, i assume, i believe, i do not believe, i doubt, i feel, i do not feel, i guess, i speculate, i think, i do not think, if anything, imo, imply, in my mind, in my opinion, in my understanding, in my view, it be possible, it look like, it do not look like, kind of, mainly, may, maybe, might, my impression be, my thinking be, my understanding be, perhaps, possibly, potentially, presumably, probably, quite, rather, relatively, seem, somehow, somewhat, sort of, supposedly, to my knowledge, virtually, would. \textbf{Boosters (score=-1):} be definite, definitely, directly, enormously, entirely, evidently, exactly, explicitly, extremely, fundamentally, greatly, highly, in fact, incredibly, indeed, inevitably, intrinsically, invariably, literally, necessarily, no way, be obvious, obviously, perfectly, precisely, really, be self-evident, be sure, surely, totally, truly, be unambiguous, unambiguously, be undeniable, undeniably, undoubtedly, be unquestionable, unquestionably, very, wholly \cite{Hyland:2005wh,URL1,URL2} \\
        Qualification & \textbf{Qualifiers (score=1):} a bit, a few, a large amount of, a little, a lot of, a number of, almost, approximately, except, generally, if, in general, largely, likely, lots of, majority of, many, more or less, most, mostly, much, nearly, normally, occasionally, often, overall, partly, plenty of, rarely, roughly, several, some, sometimes, tend, ton of, tons of, typically, unless, unlikely, usually. \textbf{Generality words (score=-1):} all, always, every, everybody, everyone, everything, never, no, no one, nobody, none, neither, not any, ever, forever \cite{Hyland:2005wh,URL2,URL3}  \\
        Arousal & \newcite{Warriner:2013gv} \\
        Dominance & \newcite{Warriner:2013gv} \\
        \bottomrule 
    \end{tabularx}
    \caption{Lexicons and regular expressions used in feature extraction.}
    \label{tab:feature_lexicons}
\end{table}

\newpage
\section{Statistical Model for Feature Effects\label{app:stat_models}}
For each feature, we use the following logistic regression model:
\begin{align*}
    \log \frac{\mathbb{P}(\mathbf{Y} = 1)}{1 - \mathbb{P}(\mathbf{Y} = 1)} = & \beta_0 + \beta_X\mathbf{X} + \alpha_1\mathbf{D}_1 + \cdots + \alpha_{|D|}\mathbf{D}_{|D|},
\end{align*}
where $\mathbf{X}$ is a continuous or binary explanatory variable that takes the value of a characteristic that we are interested in. $\mathbf{D}_d$ ($d = 1, \cdots, |D|$) is a binary variable that takes 1 if the sentence belongs to the $d$-th domain. $\mathbf{Y}$ is a binary response variable that takes 1 if the sentence is attacked or if the sentence is attacked successfully. $\beta_X$ is the regression coefficient of the characteristic $\mathbf{X}$, which is the main value of our interest for examining the association between the characteristic and the response; $\exp{\left(\beta_X\right)}$ is the \textit{odds ratio} (OR) that is interpreted as the change of odds (i.e., the ratio of the probability that a sentence is (successfully) attacked to the probability that a sentence is not (successfully) attacked) when the value of the characteristic increases by one unit. If $\beta_{X}$ is significant, we can infer that $\mathbf{X}$ has an effect on $\mathbf{Y}$. If $\beta_X$ is positive (and significant), we can infer that the characteristic and the response have positive association, and vice versa.

\newpage
\section{Important $n$-gram Features\label{app:ngrams}}
Table \ref{tab:ngrams_effects_full} shows the top 100 $n$-grams that have the highest or lowest weights for attacked sentences (vs. unattacked sentences) and for successfully attacked sentences (vs. unsuccessfully attacked).
\begin{table*}[!hp]
    \small
    \centering
    \begin{tabularx}{\linewidth}{lCC} \toprule
         & Attacked (vs. Unattacked) & Attacked Successfully (vs. Unsuccessfully)  \\ \cmidrule(r){1-2} \cmidrule(l){3-3}
        High & is are no - ? life why women should to society men a nothing 1\_) would money if\_i they n't people if * someone 2\_. human never believe 2\_) 3\_. your i\_believe and 5\_. americans tax 4 , being :\_- :\_* feel because *\_the than could republicans do be government ) sex 3\_) nobody why\_should the\_government "\_i seems religion their ca ca\_n't less 4\_. pay world war an )\_the 6\_. without ,\_why science 4\_) reason humans animals racism military selfish racist of when social 3 gun makes you speech climate get kids have can white should\_i ,\_is *\_** proven how\_can
        & is without are ? would public life women ~ weapons data how\_can usa no should if sex of\_. ,\_would n't why money \% someone the\_us customers coffee since 1\_: skills are\_a end 3\_. available ,\_they technology 2\_. - ,\_if people\_with cost need a car the pretty\_much racist so\_many to\_know third such\_as white dog could\_be towards\_the americans song actions seems formal ,\_he gender\_is nothing this\_: power see teams job years videos rates why\_would cream expectations ca god people feet global i\_believe sounds n't\_the 100 think\_that\_it crime to\_pay firstly because ,\_why immoral and\_not can\_also scooby "\_i issues \%\_of ca\_n't marriages ability in\_many \\
        Low & edit cmv i /\_? / thanks ( edit\_: [ ! post ] ]\_( this thank thank\_you comments please view | \&gt; discussion here topic sorry changed my\_view some cmv\_. posts .\_" my delta comment i\_will points responses :\_1\_. of\_you /\_) article title i\_'ll 'll = thanks\_for now 'm \&amp; got i\_'m was **\_edit above recently reddit view\_. lot i\_was below change\_my hi ’s a\_few edit\_2 on\_this again “ )\_. my\_view\_. this\_post discuss arguments you\_all deltas few there\_are 1\_. i\_'ve /\_)\_| i\_have currently edit\_2\_: comments\_. let\_me a\_lot hello let i\_still here\_. background course )\_| context you\_guys appreciate thread perspective and\_i posted
        & edit cmv i thanks / edit\_: view this thank ! 1\_. definitely ] post discussion thank\_you some 's\_a changed that\_this here i\_have tv points today responses above ,\_it\_'s ]\_( perspective both thought i\_was to\_any do\_this ( there\_are \&gt; continue\_to currently :\_i delta comments certainly taxes my you\_can discuss matters person\_a please let\_me got ,\_that not\_all 'm i\_'m more\_of n't\_want\_to obvious posts friends has\_been honest true\_. background great hypocritical case\_. work\_, account not\_the results article bit all\_the that\_would\_be grow whose thread fine\_. point\_. do\_you remember still hope now standard thanks\_for asking try\_to go started wealth = bitcoin series arguments super does\_n’t \\
        \bottomrule 
    \end{tabularx}
    \caption{$n$-grams ($n=1, 2, 3$) with the highest/lowest weights. Different $n$-grams are split by a space, and words within an $n$-gram are split by ``\_''.}
    \label{tab:ngrams_effects_full}
\end{table*}

\newpage

\section{Reproducibility Checklist\label{sec:reproducibility}}
\begin{table}[!hp]
    \centering
    \small
    \begin{tabularx}{\linewidth}{p{5cm}CC} 
        \toprule
        Criterion & LR & BERT \\
        \midrule
        Computing infrastructure & Intel(R) Core(TM) i7-3770K CPU @ 3.50GHz / 19GiB System memory & Intel(R) Core(TM) i7-8700K CPU @ 3.70GHz / 31GiB System memory / NVIDIA GP102 [TITAN Xp] \\
        Average runtime & Attacked: 225.9 mins / Successful: 31.5 mins & Attacked: 279.5 mins / Successful: 43.4 mins \\
        Number of parameters & 20,105 & 108M \\
        Validation performance & Attacked: P@1=47.4, A@3=75.8, AUC=61.8 / Successful: P@1=26.5, A@3=54.6, AUC=60.1 & Attacked: P@1=50.3, A@3=77.6, AUC=64.6 / Successful: P@1=28.3, A@3=57.2, AUC=62.0 \\
        \midrule
        Bounds for hyperparameters & Norm: \{L1, L2\} / Regularization weight: \{1e-4, 1e-3, 1e-2, 1e-1\}  & Learning rate: 1e-5 / Adam $\epsilon$: 1e-8 \\
        Hyperparameter configurations for best-performing models & Norm: L2 / Regularization weight: 1e-1 & Learning rate: 1e-5 / Adam $\epsilon$: 1e-8 \\
        Number of hyperparameter search trials & 8 & (No hyperparameter search) \\
        Method of choosing hyperparameter values &  Grid search & (No hyperparameter search) \\
        Criterion for selecting optimal hyperparameter values & AUC & (No hyperparameter search) \\
        \bottomrule
    \end{tabularx}
    \caption{Reproducibility checklist.}
    \label{tab:reproducibility_checklist}
\end{table}

\newpage

\section{Prediction Results\label{app:pred_results}}
Table \ref{tab:pred_acc_full} shows the prediction accuracy with an additional metric mean average precision (MAP). 

\begin{table*}[!hp]
    \small
    \centering
    \begin{tabularx}{\linewidth}{lCCCCCCCC} \toprule
         & \multicolumn{4}{c}{Attacked} & \multicolumn{4}{c}{Successfully Attacked} \\ \cmidrule(lr){2-5} \cmidrule(lr){6-9}
         & P@1 & Any@3 & MAP & AUC & P@1 & Any@3 & MAP & AUC \\ \cmidrule(r){1-5} \cmidrule(l){6-9}
        Random & 35.9 & 66.0 & 48.0 & 50.1 & 18.9 & 45.0 & 34.0 & 50.1 \\
        Length & 42.9 & 73.7 & 53.7 & 54.5 & 22.3 & 52.1 & 38.8 & 55.7 \\
        \cmidrule(r){1-5} \cmidrule(l){6-9}
        Logistic Regression & 47.1 & 76.2 & 56.5 & 61.7 & 24.2 & 54.5 & 41.0 & 59.3 \\
        ~($\times$) Content & 45.2 & 74.4 & 54.7 & 58.1 & 24.0 & 52.6 & 39.9 & 57.0 \\
        ~($\times$) Knowledge & 47.0 & 76.0 & 56.4 & 61.7 & 24.1 & 54.3 & 40.5 & 59.0 \\
        ~($\times$) Prop Types & 46.7 & 75.9 & 56.2 & 61.5 & 24.4 & 53.6 & 40.7 & 59.0 \\
        ~($\times$) Tone & 47.0 & 76.0 & 56.4 & 61.9 & 25.2 & 56.2 & 41.4 & 59.4 \\
        BERT & 49.6 & 77.8 & 57.9 & 64.4 & 28.3 & 57.2 & 43.1 & 62.0 \\
        \cmidrule(r){1-5} \cmidrule(l){6-9}
        Human & 51.7 & 80.1 & -- & -- & 27.8 & 54.2 & -- & -- \\
        \bottomrule
    \end{tabularx}
    \caption{Prediction accuracy.}
    \label{tab:pred_acc_full}
\end{table*}

\newpage
\section{Visualization Examples\label{app:visualization_examples}}

\begin{figure}[!hp]
    \centering
    \begin{subfigure}[t]{.6\linewidth}
        \includegraphics[width=\linewidth]{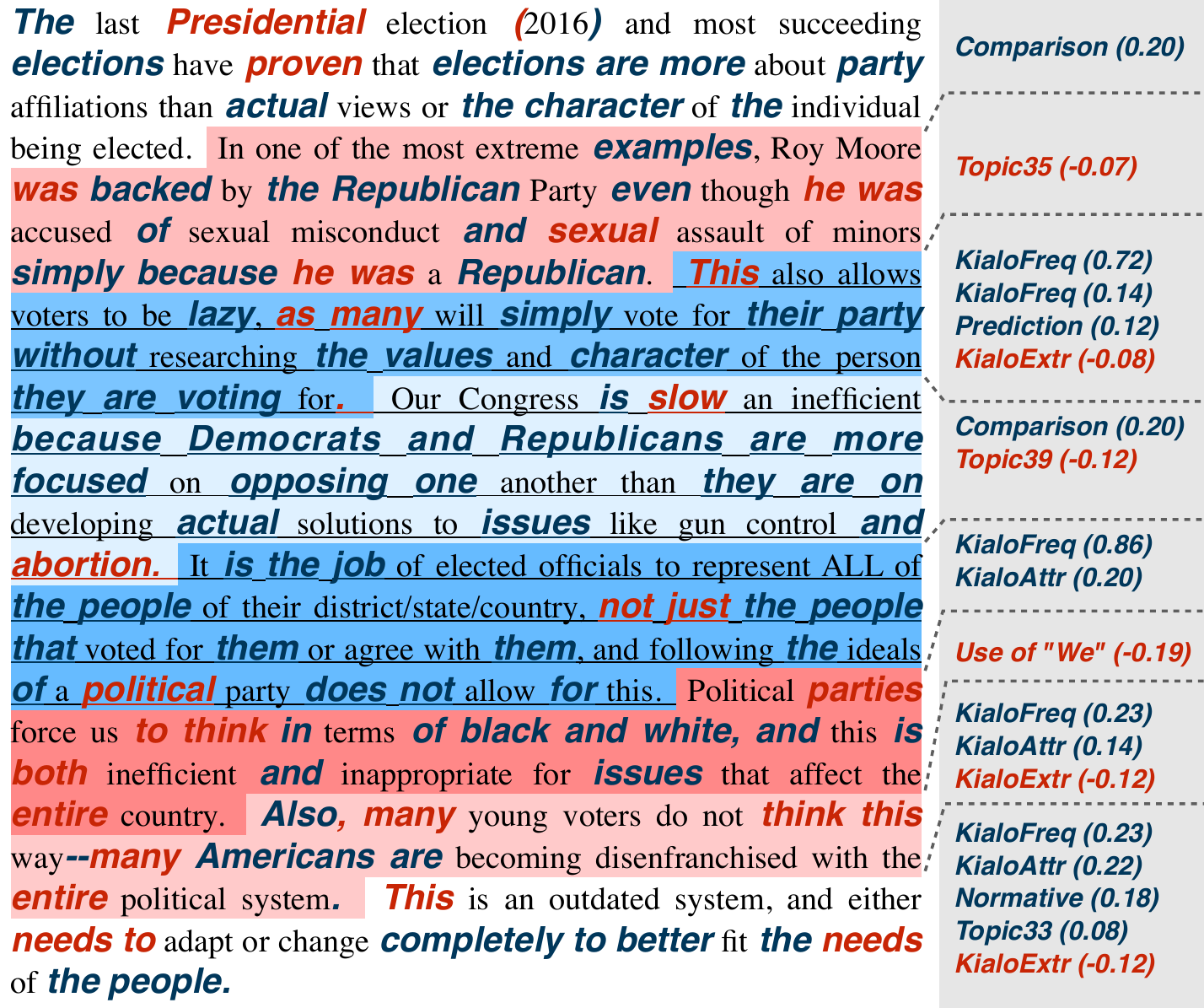} %
        \caption{Successful example 2.}
        \label{fig:pred_success2}
    \end{subfigure}
    \begin{subfigure}[t]{.6\linewidth}
        \includegraphics[width=\linewidth]{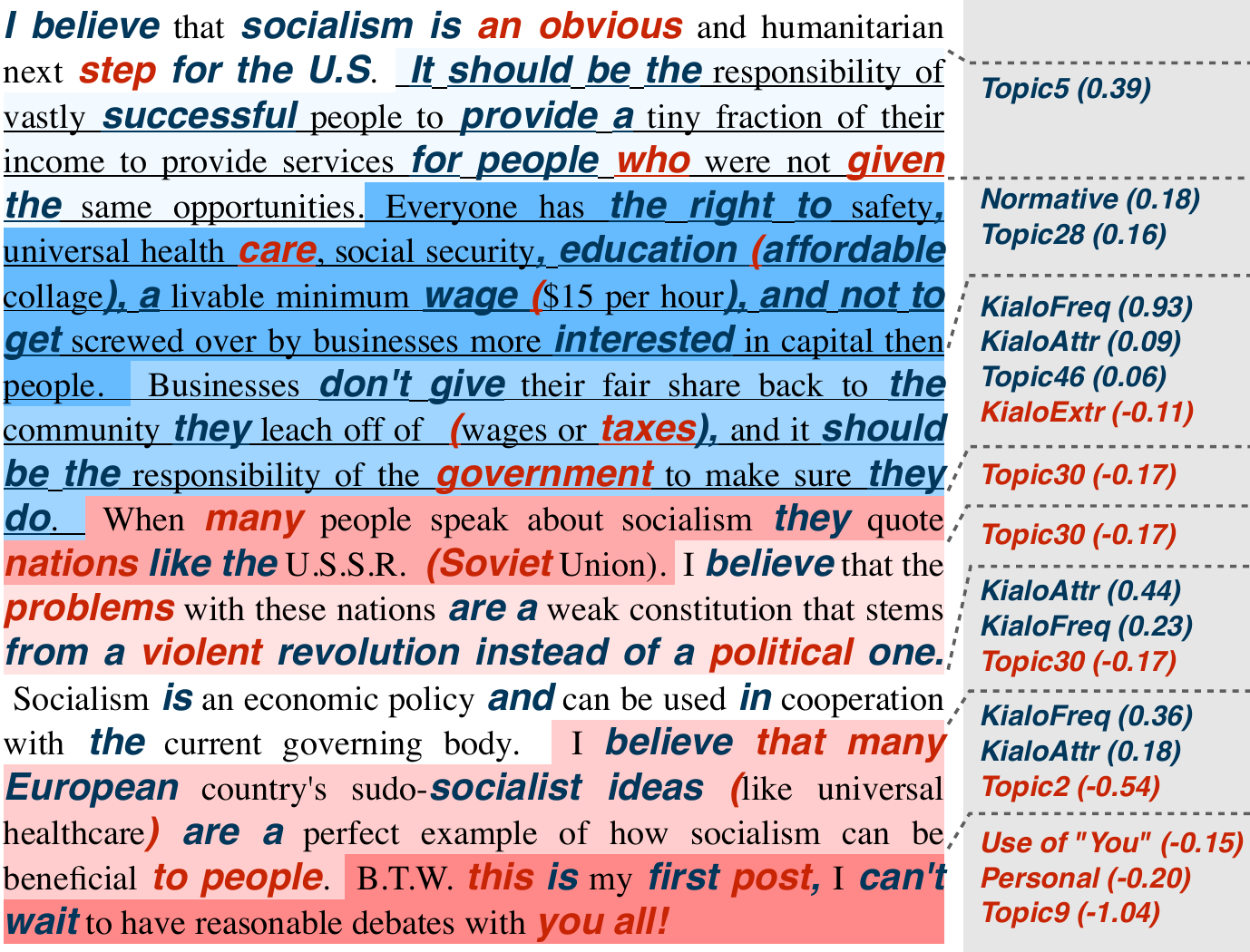} %
        \caption{Successful example 3.}
        \label{fig:pred_success3}
    \end{subfigure}
    \caption{Successful examples.}
    \label{fig:pred_success}

\end{figure}

For the successful example in Figure \ref{fig:pred_success2}, the model finds evidence for the successfully attacked sentences 3 and 5 from the external knowledge source (Kialo). Although some of the other sentences (7--8) also match Kialo statements, the degree of match is relatively low, and the model determines that their $n$-grams reduce attackability (\lex{many}, \lex{think}, \lex{needs}). Sentence 4 is properly found to have high attackability, since it makes a comparison and contains many $n$-grams predictive of attackability (\lex{because}, \lex{Democrats}, \lex{Republicans}, \lex{opposing}).

For the successful example in Figure \ref{fig:pred_success3}, topics play important roles for determining attackability. The topics of the successfully attacked sentences 2--4 all increase attackability, whereas the topics of other sentences 5--9 reduce attackability.

\begin{figure}[!hp]
    \centering
    \begin{subfigure}[t]{.6\linewidth}
        \includegraphics[width=\linewidth]{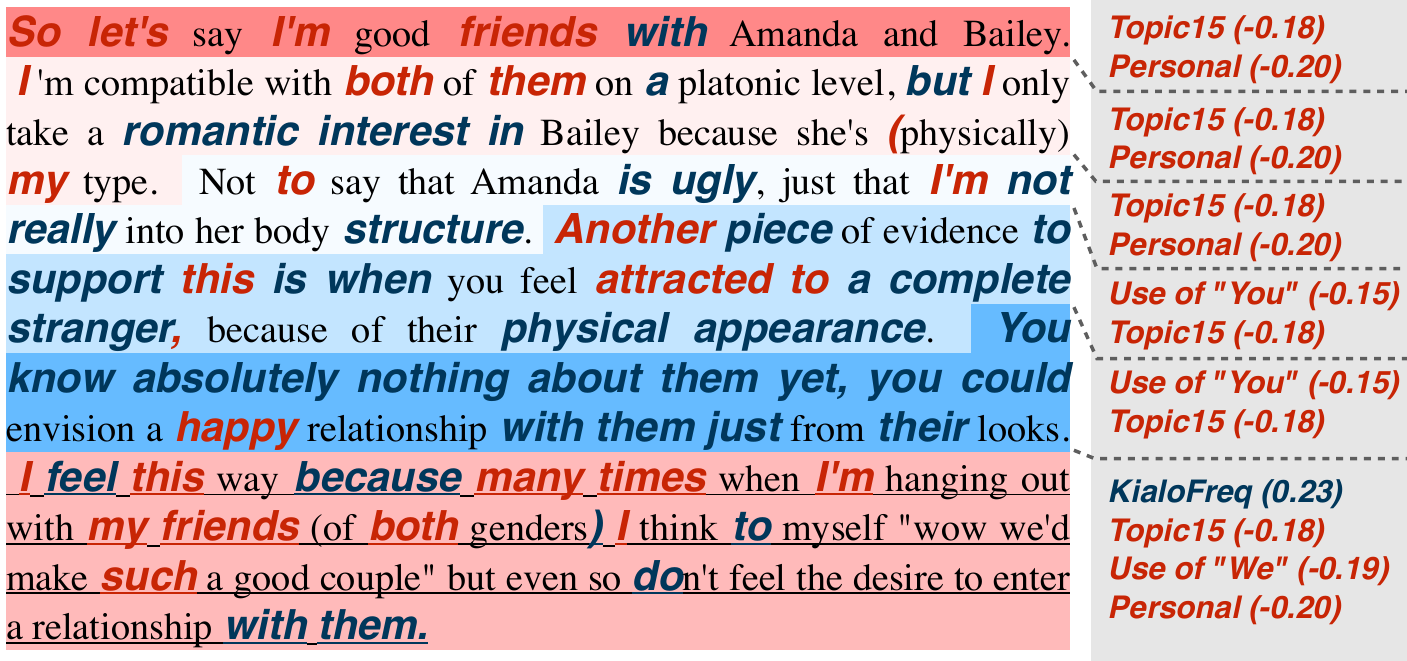} 
        \caption{Erroneous example 1.}
        \label{fig:pred_unsuccess1}
    \end{subfigure}
    \begin{subfigure}[t]{.6\linewidth}
        \includegraphics[width=\linewidth]{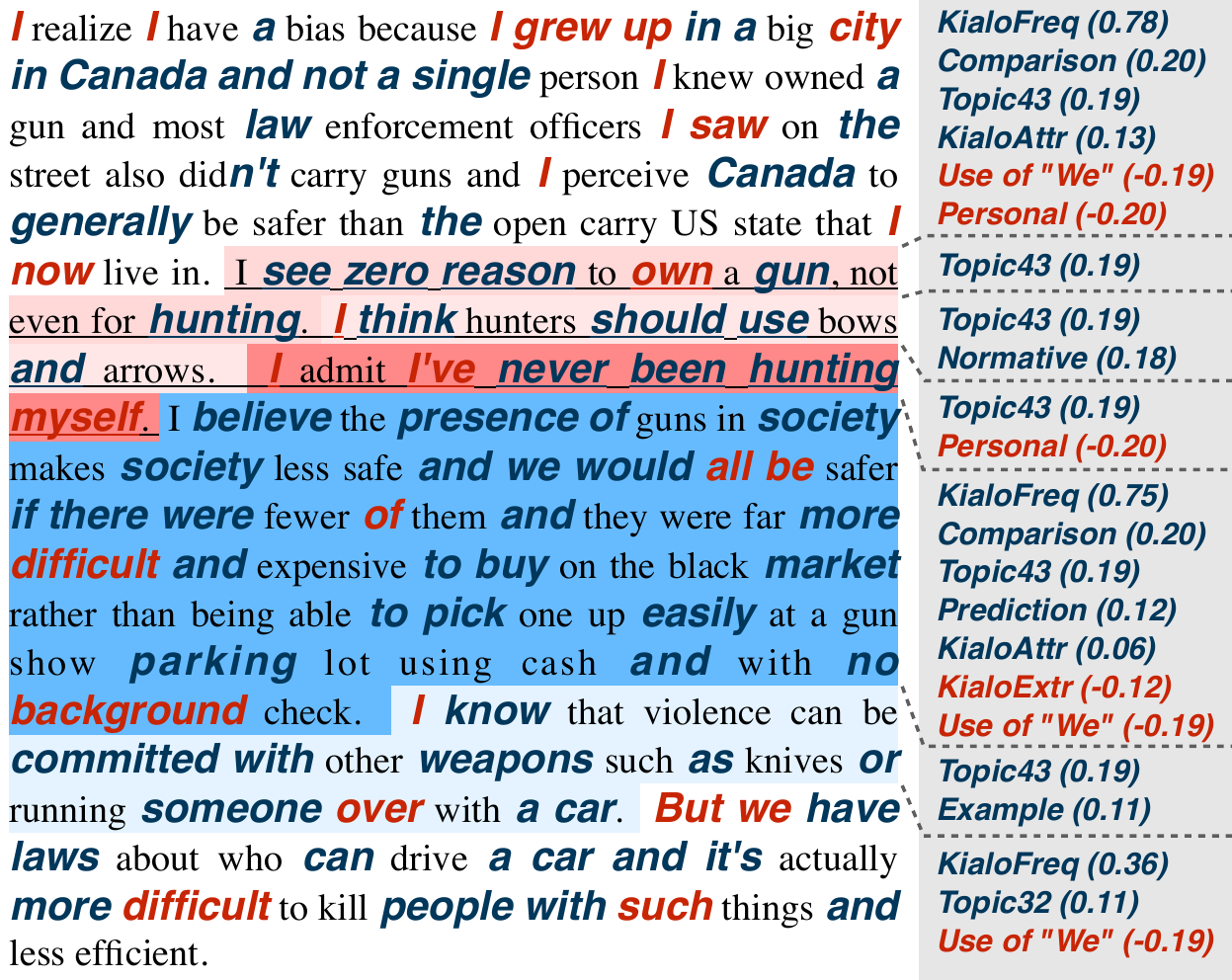} 
        \caption{Erroneous example 2.}
        \label{fig:pred_unsuccess2}
    \end{subfigure}
    \caption{Erroneous examples.}
    \label{fig:pred_unsuccess}

\end{figure}

For the erroneous example in Figure \ref{fig:pred_unsuccess1}, all sentences have relatively little evidence for attackability/unattackability. The model determines sentence 5 to have relatively high attackability because of many $n$-grams that increase attackability (\lex{know}, \lex{absolutely}, \lex{nothing}). On the other hand, the successfully attacked sentence 6 is assigned a low attackability score despite its match with Kialo statements, because its use of \lex{we}, personal stories, and certain $n$-grams (\lex{many}, \lex{times}, and \lex{friends}).

For the erroneous example in Figure \ref{fig:pred_unsuccess2}, the model finds sentence 4 to have high attackability because it matches with Kialo statements, makes a comparison and prediction, and certain $n$-grams (\lex{believe}, \lex{presence}, \lex{society}, \lex{market}). Sentence 5 is also assigned a relatively high attackability score due to its use of examples and certain $n$-grams (\lex{know}, \lex{committed}, \lex{weapons}). However, these sentences were not successfully attacked. In contrast, the successfully attacked sentences 2--4 do not have strong enough evidence for attackability compared to their negatively signals, such as personal stories and $n$-grams \lex{own} and \lex{I}.

\end{document}